\documentclass{article}
\usepackage{amsmath}
\usepackage{mdframed}
\usepackage[utf8]{inputenc}
\usepackage{textcomp}
\usepackage{amssymb}

\usepackage[numbers, square]{natbib}

\usepackage{graphicx}
\usepackage[final]{AdvML_Frontiers_2024}

\usepackage[utf8]{inputenc} % allow utf-8 input
\usepackage[T1]{fontenc}    % use 8-bit T1 fonts
\usepackage{hyperref}       % hyperlinks
\usepackage{url}            % simple URL typesetting
\usepackage{booktabs}       % professional-quality tables
\usepackage{amsfonts}       % blackboard math symbols
\usepackage{nicefrac}       % compact symbols for 1/2, etc.
\usepackage{microtype}      % microtypography
\usepackage{xcolor}         % colors
\title{Rethinking Randomized Smoothing from the Perspective of Scalability}

\author{%
  Anupriya Kumari\footnotemark[1] \\
  ECE Dept. \\
  IIT Roorkee \\
  \texttt{anupriya\_k@ece.iitr.ac.in} \\
  \And
  Devansh Bhardwaj\footnotemark[1] \\
  ECE Dept. \\
  IIT Roorkee \\
  \texttt{d\_bhardwaj@ece.iitr.ac.in} \\
  \And
  Sukrit Jindal\thanks{Equal Contribution}\\
  MFS DSAI\\
  IIT Roorkee\\
  \texttt{sukrit\_j@mfs.iitr.ac.in} \\
}

\begin{document}

\maketitle

\begin{abstract}
Machine learning models have demonstrated remarkable success across diverse domains but remain vulnerable to adversarial attacks. Empirical defense mechanisms often fail, as new attacks constantly emerge, rendering existing defenses obsolete, shifting the focus to certification-based defenses. Randomized smoothing has emerged as a promising technique among notable advancements. This study reviews the theoretical foundations and empirical effectiveness of randomized smoothing and its derivatives in verifying machine learning classifiers from a perspective of scalability. We provide an in-depth exploration of the fundamental concepts underlying randomized smoothing, highlighting its theoretical guarantees in certifying robustness against adversarial perturbations and discuss the challenges of existing methodologies.
\end{abstract}

\section{Introduction}

Machine learning (ML) has advanced significantly, notably since the development of deep neural networks (DNNs) (\cite{b84}). However, a significant challenge is the susceptibility of state-of-the-art networks to adversarial examples, which limits their application in areas of critical safety (\cite{b70, b73, b82}).
Adversarial attacks include techniques such as DeepFool, AutoAttack, and patch-based attacks (\cite{b100, b101, b102}). Existing defenses against adversarial attacks have proven ineffective against stronger methods, creating a constant cat-and-mouse game between attackers and defenders where attackers continually devise stronger attacks and defenders attempt to provide robustness against these attacks (\cite{b86, b111}). Empirical and heuristic defenses include denoising auto-encoders, using training-based defenses such as adversarial training, and defensive distillation (\cite{b103,b104,b105}). This ongoing struggle has led researchers to explore certified robustness as an alternative. In the field of certified defense, randomized smoothing has emerged as a powerful certification-based defense against adversarial attacks.

\subsection{Randomized Smoothing}
Randomized smoothing was introduced as a certification-based defense technique by \cite{b1}. The fundamental idea behind randomized smoothing is to create a smoothed classifier by applying a convolution with Gaussian noise to the base classifier. Given a base classifier $f$, we create a smoothed version of this classifier, denoted as $g$, which predicts the most likely class $c$ that the base classifier $f$ will predict for a noisy version of the input $x + \epsilon$, where $\epsilon \sim \mathcal{N}(0, \sigma^2I)$. Thus, $g$ can be calculated as:
\begin{equation} \label{eq1}
    g(x) = \arg\max_{c \in [C]} P(f(x+\epsilon) = c) \text{, where } \epsilon \sim \mathcal{N}(0, \sigma^2 I)
\end{equation}

While \cite{b77} and \cite{b2} showed that the smoothed classifier $g$ will consistently classify within a certified radius around the input $x$ under $l_2$ norm considerations, \cite{b1} was the first paper to demonstrate tight robustness guarantees for a randomized smoothing classifier against adversarial attacks constrained under the norm $l_2$. For empirical calculation they made use of Clopper-Pearson bounds based on the Neyman-Pearson Lemma by making use of Monte Carlo Sampling. 

% Since Cohen et al.'s seminal work, there has been a large number of works which have been published with regards to randomized smoothing. The organization of the paper is structured into two main sections. Section I critically examines the major issues associated with randomized smoothing, categorizing these challenges into broad domains. Subsequently, a detailed discussion ensues on these challenges, followed by an exploration of future research directions.

% Moving to Section II, we delve into significant advancements that focus on enhancing radius certifications beyond those outlined in Cohen et al. . This section also addresses the challenges presented in Section I, providing discussions for the same. The appendix discusses applications of randomized smoothing, which extend beyond the original focus of Cohen et al., from the lens of scalability.

% \subsection{Rethinking Scalability in Randomized Smoothing: An Analytical Perspective}

Since Cohen et al.'s seminal work, there has been a large number of works which have been published with regards to randomized smoothing. Due to its theoretical guarantees, randomized smoothing is commonly hailed as a crucial certification-based defense mechanism. However, in our work through analysis, we raise a fundamental question:

\textbf{\textit{Despite all the theoretical assurances, is randomized smoothing truly scalable?}}

Throughout our examination, we highlight the limitations of existing works that predominantly emphasize theoretical guarantees while overlooking the critical aspect of computational complexity, especially during inference, and dimensionality considerations. This concern holds significant weight, particularly for machine learning applications requiring minimal inference time, where randomized smoothing may fall short as a practical solution. Hence, we scrutinize each work from the perspective of scalability.

The paper is structured as follows: We begin with a comprehensive overview of the various works in the context of randomized smoothing, categorizing them into two broad areas: "Challenges of Randomized Smoothing" and "Improving Randomized Smoothing". Sections 2 and 3 delve into these categories, respectively, providing an in-depth discussion of each. Within these sections themselves we have scrutinized the works from the perspective of scalability. In the appendix we have included more detailed discussions of the works in randomized smoothing related to section 2 and section 3. In Section 4, we revisit our discussions in the preceding sections, and discuss the aspect of scalability from a real-world perspective, and provide some future directions and challenges associated with them that the community can focus as a whole. 

\begin{figure}
    \centering
    \includegraphics[width=1\linewidth]{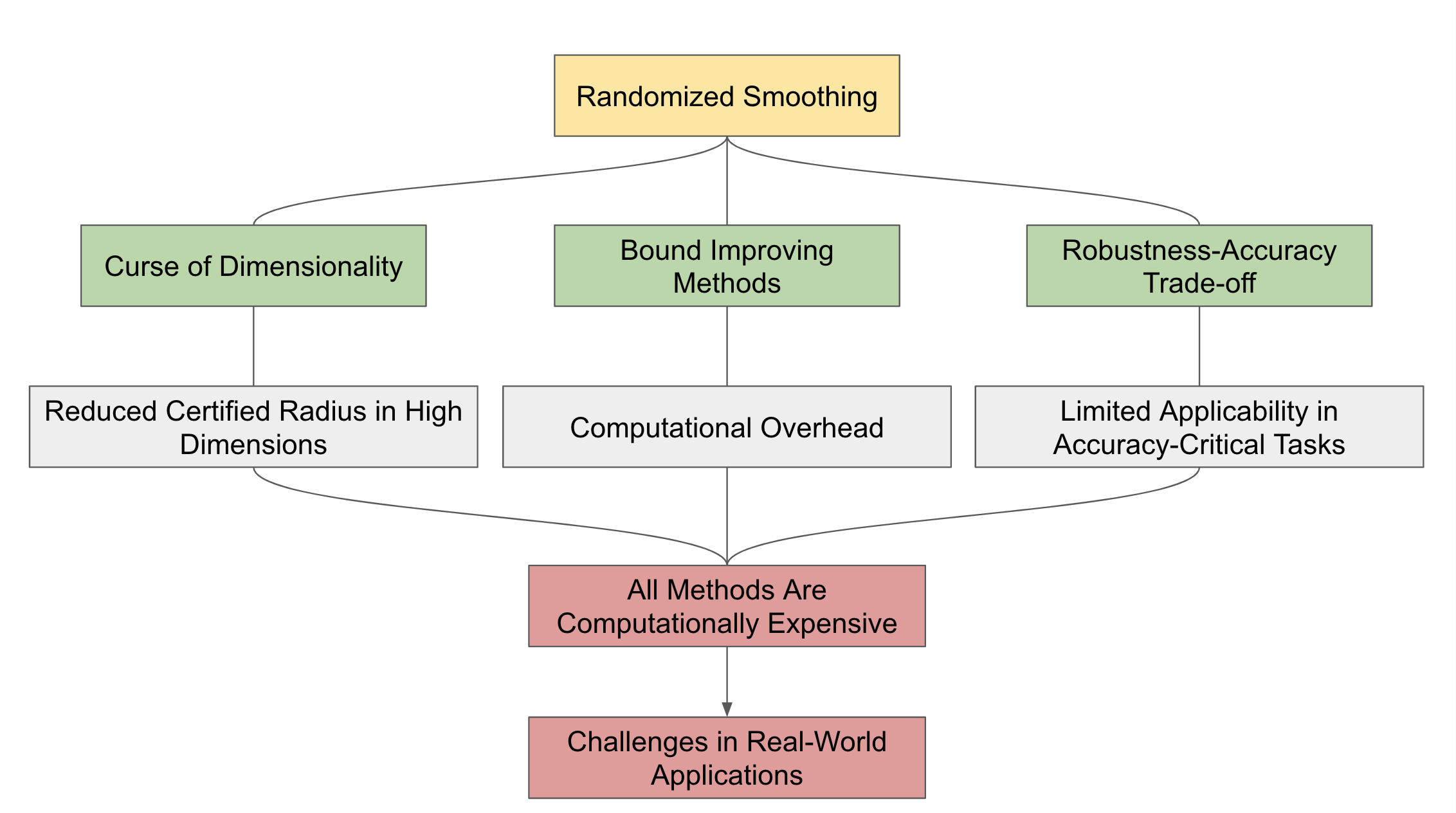}
    \caption{This flowchart provides an overview of the main scalability challenges with RS.}
    \label{fig:enter-label}
\end{figure}

\section{Challenges of Randomized Smoothing}

\subsection{Curse of Dimensionality}

In their work, \cite{b1} left the case of general $l_p$ norms as an open problem and only discussed randomized smoothing in the context of $l_2$ norms. Additionally, they state that, despite the certified radius being independent of the input dimension $d$, randomized smoothing can be easily scaled to higher dimension images as in the case of higher dimension images, we can increase the variance of the noise, which will eventually lead to increase in the robust radius. Later, \cite{b8} and \cite{b13} showed that as the dimension of the input increases, it becomes increasingly difficult to defend against the general $l_p$ norm cases for $p>2$ as the certified radius decreases. Particularly, for the case of $l_\infty$, which is a common norm used in empirical attacks, the certified radius is equal to the certified radius for the $l_2$ norm, divided by $\sqrt{d}$. In general, there is a tight $O\left(\min\left(1,d^{\frac{1}{2}-\frac{1}{p}}\right)\right)$ dimensionality bound for $l_p$ norms without using any additional information other than the Neyman-Pearson technique, which cannot be improved on otherwise (\cite{b71}).

We note that there has been an in-depth theoretical exploration of the issue termed the curse of dimensionality, but little work to address the issue itself. In this regards, we make note of two specific works. \cite{b44} implicitly utilise the manifold hypothesis and show that using various dimensionality reduction techniques improves robust guarantees in theory, as opposed to heuristic approaches mentioned in previous works. \cite{b3} have provided proofs and a possible solution to the issue of the curse of dimensionality using a technique called double sampling randomized smoothing (DSRS). They utilize a primary distribution in the form of a general Gaussian distribution and additional information in the form of a sampled probability from a secondary truncated distribution and, they provide a tighter radius for the $l_2$ and prove that their method can certify $\Theta\left(\sqrt{d}\right)$ robust radius bounds. Among these, we note that the former work can address the computational cost issue as well by working with the data in a lower dimension, but the latter introduces a computational overhead instead.

\subsection{Robustness-Accuracy Trade-off}
In adversarial robustness literature, not just limited to RS, there is a focus on increasing the robustness bounds of a classifier. While this seems to be an optimal goal, there has been literature suggesting that increasing robustness comes with a trade-off in accuracy (\cite{b78, b83}). Both empirically and theoretically, it is seen that robustness and accuracy exist in a trade-off, contrary to what one would expect intuitively. Even in the field of RS, there have been works which have analysed the theoretical justification of the robustness-accuracy trade-off (\cite{b54, b36, b35}). Intuitively seen, randomized smoothing involves smoothing, the base classifier with Gaussian noise. While this provides certified robustness, it can be inferred that there is some loss in accuracy due to noisy inputs. Specifically, increasing the noise hyper-parameter $\sigma$ results in a larger robustly certified radius but decreased accuracy of the model. This challenge particularly limits the applicability of RS in accuracy-oriented tasks.

While as noted above, there is not a dearth of works which study the theoretical robustness-accuracy trade-off, there are very few works which aim to actually address it. In this regard, we note the work on a compositional architecture termed ACES (\cite{b52}) which decides on a per-sample basis whether to use the robust-but-less-accurate smoothed model or the accurate-but-less-robust base classifier for prediction using a learnable selection algorithm that can again be made robust by further smoothing. This is done in the light of optimising randomized smoothing specific robustness and accuracy using compositional architectures, as introduced and proposed earlier in \cite{b98}.

\subsection{High Inference Cost}

We note that high inference cost is an issue which plagues randomized smoothing. This is because it utilizes Monte Carlo sampling, which requires multiple passes of the noisy samples through the model, leading to high inference costs and for significant radius sizes, typically $10^5$ or more samples are required as per \cite{b1}.

A controlled way to address this was to trade off the average certified radius (ACR) with the number of samples required using an input-specific scheme (\cite{b25}). Another way of reducing the computation cost could be by lowering the inference time of the smoothed model by techniques such as pruning, quantization of the base model, or knowledge transfer. But this essentially means we have changed the actual model and the theoretical guarantees may not hold. In this regard, a incremental certification technique for randomized smoothing is Incremental Randomized Smoothing (IRS) which improves performance on probabilistic techniques like randomized smoothing compared to existing deterministic techniques for incremental certification (\cite{b11}). This works by reusing information from the smoothed classifier of the base classifier to reduce computational costs. On a related note, we note Certified Robustness Transfer (CRT) as a training technique based on knowledge transfer (\cite{b124}).

While these techniques display promising empirical results, they often lack the theoretical justification behind their validity. Further, there is much still left to be explored in this regard, as can be seen by the few numbers of works written in this regard.

\subsection{Challenges in the Scalability of RS}

In this section, we explored the challenges which affect RS. While the robustness-accuracy tradeoff, as noted, is a challenge which affects the entire field of adversarial robustness and even ML as a whole, the other two, viz., the curse of dimensionality and the issue of high inference costs, expressly limit the scalability of RS. Further, on a broad level, we note these challenges are acknowledged and understood in their impact, whether implicitly, as is true for computational costs, or explicitly, as is seen for the curse of dimensionality. However, contrary to expectations, little research is devoted to developing solutions that truly address these issues and improve the scalability of RS.

\section{Improving Randomized Smoothing}

Randomized smoothing offers a tight lower bound on the certified radius for any classifier. As with any certified robustness technique, these bounds are lower than the upper bounds arising from empirical verification techniques, which are based on heuristics and empirical evidence instead of theoretical guarantees. Thus, there has been much research into improving the performance of randomized smoothing and improving its average certified radii. Further, there is a need to address the challenges faced in RS and develop various techniques. In this section, we explore specific optimisations that improve the robustness certificates given by RS and solve various issues encountered in RS.

\subsection{Randomized Smoothing for Training}

The idea of incorporating randomized smoothing objectives in adversarial training roots from the intuition that doing so will result in boosted performance and improved robustness. Further, addressing the challenge of training a good base classifier that is accurate and robust when smoothed using a soft classifier with a hard classifier led to improved results (\cite{b92}). Multiple works in this regard achieve larger average certified radii in less training time than state-of-the-art adversarial training algorithms (\cite{b15, b37, b112}). Further, these methods may use adaptive techniques and improve on the tightness of the robust bounds. These approaches not only are they computationally efficient and empirically sound, but also orthogonal to other empirical defense schemes. We can link these results to scalability and conclude that these methods make RS scalable).

\subsection{Improving Performance of Pre-Trained Classifiers}
Improving the performance of pre-trained classifiers by turning them from non-robust into provably robust ones is desired to maintain their confidentiality, integrity, and availability. This includes using computationally inexpensive denoisers as denoised smoothing can be applied to both white and black box settings, making it the state-of-the-art approach to defending pre-trained classifiers with improved methodologies (\cite{b20, b57, b49}). While the work so far in this domain seems somewhat promising, there remains scope for future research to further improve the certified accuracy of high-dimensional datasets with increasing $\sigma$.

\subsection{Leveraging Local Information}
The original randomized smoothing technique makes use of Monte-Carlo sampling to derive certified radii. This information is fully summarised for an input $x$ by $p_A(x)$, the probability of the top class, and does not require any other information beyond the standard deviation $\sigma$ of the Gaussian. There has been a series of works which explore the utilisation of the gradient information, for second-order smoothing, or higher-order derivatives to achieve tight bounds, addressing the curse of dimensionality (\cite{b116, b22, b26}).

While these works explore optimization using local information and achieve improved bounds, a drawback of this approach is that the computational overheads associated with calculating local information add heavily to the already expensive smoothing convolution with the gains they add being bound (\cite{b16}).

\subsection{Ensemble-Based Methods}

Randomized smoothing bounds can be improved using ensemble-based methods. These methods work by aggregating the results of multiple classifiers, somewhat similar to how RS involves smoothing the base classifier with multiple noisy inputs and using those results to create the base classifier.

Various works have shown state-of-the-art results using ensembles as the base model for RS. One insight offered in the work is that the reduced variance of ensembles over the perturbations introduced in RS leads to significantly more consistent classifiers, while still maintaining robustness (\cite{b60, b96}). Further, these papers address the inference cost issue by developing techniques to reduce the prediction and certification cost of the models, such as patch-based sampling or adaptive prediction algorithms or reducing sample complexity with data-dependent adaptive sampling (\cite{b29, b48}).

Despite the prohibitive costs of ensemble training, adaptive algorithms have been shown to reduce the computational overhead. Overall, ensemble methods provide a satisfactory approach to improve robustness guarantees with reduced costs as compared to other similarly expensive approaches.

\subsection{Input-Specific Information}
RS uses a data-independent hyperparameter, $\sigma$, for the Gaussian noise $\epsilon \sim \mathcal{N}(0, \sigma^2I)$ to be convoluted with the base classifier. Even though this forms a basis of the robustness-accuracy tradeoff, it is also important to note that the issue of an optimal choice of $\sigma$ is important to improve certified robustness using TS. To deal with this, \cite{b17} propose the use of additional information which can be extracted from the data itself which improves the choice of the hyperparameter $\sigma$ to achieve a larger certified radius. Related to this, there have been a variety of works which gradually change the structure of the covariance matrix used in RS, making it anisotropic and data-dependent and rotation-robust (\cite{b17,b47,b63,b41,b120}), albeit without theoretical guarantees and while suffering from the curse of dimensionality (\cite{b24}). Further, many techniques use information beyond the Neyman-Pearson lemma, such as classifier-specific information, data-dependent information, local geometric information and non-Gaussian distributions for noise convolution (\cite{b5,b25,b24,b39}). However, these methods only add a computational overhead, instead of working towards reducing the inference costs. In this, we point out the work on QCRS (\cite{b62}) which as a computationally efficient approach with less overhead and is one of the few input-specific methods to address the complexity issue directly.

\subsection{Improving Randomized Smoothing for Scalability} 

Throughout our examination in this section, we observe that most new optimizations in randomized smoothing do not directly address the curse of dimensionality or the issue of high inference costs. Instead, they tend to introduce new techniques, each with its own computational overhead and dimensionality challenges, and then propose solutions for that additional cost. One such example is that of input-dependent techniques, which add input-specific components over an already costly randomized smoothing process, as pointed out in that specific subsection. While this certainly helps improve the applicability of a technique individually, it does not enhance the overall scalability of randomized smoothing, hence this foundational issue persists.

\section{Scalability in Practice}

When considering the practical scalability of RS, several key challenges emerge across various application domains. In high-dimensional data scenarios, such as high-resolution images or complex sensor inputs, \textbf{the curse of dimensionality} significantly hampers RS's effectiveness. Although a lot of works are there that discuss this issue but as already mentioned in section 2 most of these work are theoretical in nature and only a few provide practical solutions to this issue. Even then at times those solution bring additional computational overhead.

Additionally, The \textbf{high inference cost} poses difficulties for real-time systems requiring rapid decision-making, like autonomous vehicles or live video analysis. In general, All the methods of RS that we have discussed are computationally very expensive, irrespective of the challenge or the improvement they are doing. Additionally, some methods put an additional \textbf{computational overhead} over RS as discussed in some of the sub sections of section 3. Finally comes the general issue of \textbf{robustness-accuracy trade-off}, which further limits the use of RS in domains which require high accuracy but at the same time high reliability such as medical domain.

\subsection{Future Directions for Scalable RS}

To truly make RS scalable, future research could focus on developing practical techniques with theoretical justifications focused on the following:
\begin{enumerate}
    \item Efficient dimensionality handling: Developing techniques that can effectively deal with high-dimensional inputs without significant information loss or computational overhead. 
    \item Reduced inference costs: Exploring methods to dramatically reduce the number of samples required for certification without compromising on robustness guarantees.
    \item Adaptive smoothing: Developing techniques that can dynamically adjust the level of smoothing based on the input and the required level of robustness, potentially reducing unnecessary computations.
    \item Compression techniques: Investigating ways to compress the smoothed classifier without significant loss in robustness guarantees.
\end{enumerate}
\section{Conclusion}

This paper has reviewed randomized smoothing (RS) as a certification-based defense against adversarial attacks in machine learning, focusing on its theoretical foundations, practical challenges, and recent advancements and scrutinizing them from the perspective of scalability. While RS offers robust theoretical guarantees, its practical application faces challenges like the curse of dimensionality and high computational costs, limiting its effectiveness in real-world, high-dimensional scenarios. Despite various improvements, such as training-based methods and ensemble techniques, these often worsen scalability issues.

The persistent robustness-accuracy trade-off requires careful consideration, with potential solutions like compositional architectures adding complexity. Future research should focus on enhancing RS's scalability by developing efficient techniques for high-dimensional inputs, reducing sample requirements for certification, designing hardware-aware algorithms, and exploring adaptive smoothing and compression methods. In summary, while RS provides certified robustness against adversarial attacks, its scalability remains a significant barrier to widespread adoption. Addressing these challenges is essential for bridging the gap between RS's theoretical strengths and practical deployment in large-scale applications.

\bibliography{main}

%%%%%%%%%%%%%%%%%%%%%%%%%%%%%%%%%%%%%%%%%%%%%%%%%%%%%%%%%%%%

\appendix

\section{Appendix / supplemental material}

\subsection{Curse of Dimensionality}

In their work, \cite{b1} left the case of general $l_p$ norms as an open problem, suspecting that smoothing with other noise distributions might provide robustness guarantees for these cases, and only discussed randomized smoothing in the context of $l_2$ norms. Additionally, they state that, despite the certified radius being independent of the input dimension $d$, randomized smoothing can be easily scaled to higher dimension images as in the case of higher dimension images, we can increase the variance of the noise, which will eventually lead to increase in the robust radius. This approach has its issues, as will be discussed in the next section. Later, two independent works, \cite{b8} and \cite{b13}, showed that as the dimension of the input increases, it becomes increasingly difficult to defend against the general $l_p$ norm cases for $p>2$ as the certified radius decreases.

\cite{b8} showed that, given any noise distribution $\mathcal{D}$ and the noise vector $\epsilon \sim \mathcal{D}$, 99\% pixels of the noise vector, must satisfy $\mathbb{E}\epsilon_{i}^2=\Omega(d^{1-\frac{2}{p}}r^2\frac{1-\delta}{\delta^2})$, where $r$ is the robust radius and $\delta$ is the difference between the probability scores of the top two classes predicted by the smoothed classifier (highest score-class $c_A$ and runner-up class $c_B$). They further proved that the problem of finding the certified radius in the $l_\infty$ case can be approximated as finding the $l_2$ radius times $1/\sqrt{d}$ of that radius.

Subsequently, \cite{b13} show, in the specific case of a generalized Gaussian distribution, tighter bounds than those attained by \cite{b8}. Their work states that the certified radius for an $l_p$ norm decreases as $O\left(d^{\frac{1}{2} - \frac{1}{p}}\right)$ with data dimension $d$ for $p > 2$. The family of distributions discussed in this paper includes the Laplacian, Gaussian, and uniform distributions, commonly employed in literature for randomized smoothing techniques. \cite{b80} then extend the discussion for the case of spherical symmetric distributions (which includes the Gaussian distribution) and present upper bounds for $p > 2$. It also discusses upper bounds for $p = 2$ for many popular smoothing distributions, which have yet to be discussed. \cite{b4} go into more detail about $l_1$, $l_2$, and $l_\infty$ threat models to derive from the class of i.i.d. distributions, the most optimal smoothing distribution using the concept of Wulff Crystals. Overall, their results in this regard can be summarised as follows: the uniform distribution is the most optimal choice for $l_1$ attacks, the Gaussian distribution, as mentioned by \cite{b1}, is the most optimal for $l_2$ attacks, and that for $l_\infty$, the Gaussian distribution performs best using appropriate approximations. There is a tight $O\left(\min\left(1,d^{\frac{1}{2}-\frac{1}{p}}\right)\right)$ dimensionality bound for $l_p$ norms without using any additional information other than the Neyman-Pearson technique.

\cite{b71} provide a different perspective and argue about the information-theoretic limitations faced by RS and how they are not intrinsic but a byproduct of the current certification methods. They further discuss that these certificates need to include more information about the classifier and ignore the local curvature of the decision boundary, which leads to sub-optimal robustness guarantees as the dimension of the problem increases. It is possible to overcome this by generalizing the Neyman-Pearson Lemma and collecting more information about the classifier. The paper shows that it is possible to approximate the optimal certificate with arbitrary precision by probing the decision boundary with several noise distributions. More specifically, the paper achieves this without sampling in high dimensions by combining uniform and Gaussian distribution and leveraging the isotropic properties of the latter. Further, this process retains natural accuracy as it is executed at certification time. The paper then gives theoretical insight into how to mitigate the computational cost of a classifier-specific certification.

\cite{b44} present notable work in the sense that they implicitly propose a solution to the curse of dimensionality, suggested by \cite{b8}. It combines the limitations of randomized smoothing in light of the curse of dimensionality with the manifold hypothesis (\cite{b79}) and shows that using various dimensionality reduction techniques improves robust guarantees in theory, as opposed to heuristic approaches in previous works by \cite{b81} and \cite{b82}.

So far, only \cite{b3} have provided proofs and a possible solution to the issue of the curse of dimensionality using a technique called double sampling randomized smoothing (DSRS). They utilize a primary distribution in the form of a general Gaussian distribution and additional information in the form of a sampled probability from a secondary truncated distribution that is not necessarily different from the primary distribution. For $l_2$ norm, they provide a tighter radius and prove that their method can certify $\Theta\left(\sqrt{d}\right)$ robust radius bounds. As mentioned by \cite{b1}, for the $l_\infty$ case, the certified radius is equivalent to $1/\sqrt{d}$ times the $l_2$ robust radius, which on using DSRS would remove the dependency on the dimensions of the input.

We note that there has been an in-depth theoretical exploration of the issue termed the curse of dimensionality. The base conclusion of all papers remains that there is a tight bound on certified radii that decreases with increasing dimensionality, which arises from statistical, probabilistic, and information-theoretic perspectives based on the existing methodology for RS. While we certainly encourage more research into that aspect, we notice the dire lack of solutions for the same. Apart from DSRS, no breakthrough has been achieved to solve this issue. As a result, we encourage exploring various solutions to this issue.

\subsection{Robustness-Accuracy Trade-off}

In adversarial robustness literature, not limited to RS, there is a focus on increasing the robustness bounds of a classifier. While this seems to be an optimal goal, there has been literature suggesting that increasing robustness comes with a tradeoff in accuracy (\cite{b78} and \cite{b83}). It has been studied for adversarial robustness in much detail, including theoretical considerations. There are accuracy costs of improving robustness in a model. Both empirically and theoretically, it is seen that robustness and accuracy exist in a trade-off, contrary to what one would expect intuitively. Randomized smoothing is also vulnerable to this trade-off, and there is much to explore in this regard, leaving this an important open aspect of research for randomized smoothing techniques. Intutively seen, randomized smoothing involves smoothing, the base classifier with Gaussian noise. While this provides certified robustness, it can be inferred that there is some loss in accuracy due to noisy inputs. Termed as the robustness-accuracy tradeoff, we explore this issue concerning RS in this subsection. Specifically, increasing the noise hyperparameter $\sigma$ results in a larger robustly certified radius but decreased accuracy of the model. One must note that there also exists, in general, a trade-off between accuracy and robustness in RS concerning a smaller robust radius for more accurate models, not just limited to the choice of $\sigma$.

Specific to RS, the theoretical and numerical aspects of Gaussian smoothing are explored in the work by \cite{b54}, along with which they provide theoretical results showing the limitations of RS in terms of classification accuracy. Mainly, they characterize and identify the conditions under which Gaussian smoothing leads to a decrease in classification accuracy, provide theoretical lower bounds for the magnitude of this effect, numerically inspect the behaviour of the certified radius, and use tools from information theory to analyse the effects of Gaussian smoothing during training augmentation, concluding that it leads to information loss and finally validate their results empirically.

A theoretical analysis of robustness-accuracy in terms of the benign risk in accuracy, i.e., accuracy on non-adversarially-corrupted samples, associated with applying randomized smoothing on a classifier trained using noise augmentation for the base classifier is given by \cite{b36}. This arises from the observation of improved performance when using a noise-augmented base classifier instead of one without any such augmentation. They derive an upper bound for any data distribution used for smoothing, which suggests randomized smoothing harms the classifier's accuracy. This practice has pervaded most literature with ample empirical support. The paper then suggests certain distributions where smoothing may be beneficial based on the separation regime. Also, it suggests optimisation for randomized smoothing, using different noise parameters to train the base classifier. They also discuss the theoretical intuition behind this, viz., the empirical observation that real data lies on a lower dimension than the actual data dimension.

A compositional architecture termed ACES (\cite{b52}), which aims to tackle this robustness-accuracy trade-off, decides on a per-sample basis whether to use the robust-but-less-accurate smoothed model or the accurate-but-less-robust base classifier for prediction using a learnable selection algorithm that can again be made robust by further smoothing. This approach improves the trade-off between robustness and accuracy in current models and can be used as an orthogonal optimisation to other methods. This is done in the light of optimising randomized smoothing specific robustness and accuracy using compositional architectures, as introduced and proposed earlier by \cite{b98}.

A notable work by \cite{b35} articulates and proves two major limitions regarding RS: (I) the decision boundary of the smoothed classifier will shrink, resulting in a discrepancy in class-wise accuracy, and (II) applying noise augmentation in the training process does not resolve the shrinking issue because of inconsistent learning objectives. The paper proceeds to review adversarial robustness certification, exposes the significant hidden cost of RS, which includes biased predictions using evidence from both real-life and synthetic datasets, provides a comprehensive theory exposing the root of the biased prediction-shrinking phenomenon and then discusses the effects of data augmentation on the shrinking phenomenon and its implications. In essence, they arrive at two observations - the need to limit the use of high values of smoothing factor $\sigma$ and a data geometry-dependent augmentation scheme to counteract the shrinking effect caused by smoothing properly. This challenges what we already know about randomized smoothing, which, as introduced by Cohen et al., uses data-independent noise parameters. Namely, the Gaussian noise, $\epsilon \sim \mathcal{N}(0, \sigma^2I)$ which is convoluted with the base classifier is data-independent and $\sigma$ can be thought of as a hyperparameter representing a trade-off between accuracy and robustness. Too low of a value will give smaller certified radii, which are given, after appropriate approximations, as the certified radius $r = \sigma\Phi^{-1}(\underline{p_A})$, which is directly proportional to $\sigma$. Using too large of a value for $\sigma$ will reduce the accuracy in calculating $\underline{p_A}$ due to highly noisy training inputs.

Overall, we notice ample theoretical and empirical evidence supporting the existence, and its pervasivity in most RS techniques. However, there is very little literature aimed at actually tackling this problem. We strongly suggest more works to explore solutions to this challenge and improve existing methods like ACES using newer techniques and methodology, not just for RS but also for adversarial robustness in general.

\subsection{High Inference Cost}

Randomized smoothing utilizes Monte Carlo sampling, which requires multiple passes of the noisy samples through the model, leading to high inference costs, at the same time it has been shown that certified radius increases with the no. of passes or no. of samples ($n$) and for significant radius size typically $10^5$ or more samples are required (\cite{b1}). This implies that during inference we need to do a forward pass to have a significant certified radius and essentially increases the computational cost by a large margin. 

A controlled way to trade off the average certified radius (ACR) with the number of samples required was introduced by \cite{b25}. They changed their sampling scheme from an input-agnostic scheme to an input-specific scheme. To achieve this input-specific sampling, first a relatively loose two-sided Clopper-Pearson interval is calculated for a given input using a smaller sample size. Then, a mapping of sample size to the relative decline in average radius is computed for the input. This mapping shows a trade-off between the number of samples and the average certified radius.

Another way of reducing the computation cost could be by lowering the inference time of the smoothed model by techniques such as pruning, quantization of the base model, or knowledge transfer. But this essentially means we have changed the actual model and the theoretical guarantees may not hold. \cite{b11} study this scenario and formulate an incremental certification technique for randomized smoothing called Incremental Randomized Smoothing (IRS). This improves performance on probabilistic techniques like randomized smoothing compared to existing deterministic techniques for incremental certification. In particular, incremental certification deals with certifying robustness bounds for the smoothed classifier of a given modification of a base classifier, reusing information from the smoothed classifier of the base classifier to reduce computational costs. It does so based on the disparity between the classifiers and certain approximation techniques. On a related note, \cite{b124} propose a similar knowledge transfer scheme, Certified Robustness Transfer (CRT), for randomized smoothing for training $l_2$ certifiably robust image classifiers with comparable levels of robustness using a pre-trained classifier that is certifiably robust as a teacher. Both techniques demonstrate significantly reduced costs, indicating the empirical effectiveness of their methods.

It is pretty clear from the number of works in this section that this aspect of randomized smoothing is virtually untouched and has a lot of scope for improvements and future directions.

\subsection{Randomized Smoothing for Training}

The idea of incorporating randomized smoothing objectives in adversarial training (\cite{b92}) roots from the intuition that doing so will result in boosted performance and improved robustness. Further, addressing the challenge of training a good base classifier that is accurate and robust when smoothed led to significantly better results (\cite{b37}). \cite{b92} consider a smoothed classifier which is a generalization of equation (\ref{eq2}), to soft classifiers, namely, functions $f: \mathbb{R}^d \rightarrow P(y)$ where $P(y)$  is the set of probability distributions over $y$. They then train a hard classifier which essentially takes the argmax of the soft classifier, using adversarial examples generated by describing an adversarial attack against the smoothed soft classifier (\cite{b112} and \cite{b113}). This results in a boosted empirical robustness and substantially improved certifiable robustness using the certification method of \cite{b1}. \cite{b37} improved upon this by deriving a new regularized risk in which the regularization can 'adaptively encourage' the training of the base classifier. It is computationally efficient and can be implemented in parallel with other empirical defense methods, under standard (non-adversarial) and adversarial training schemes. They also designed a new certification algorithm, T-CERTIFY, that can leverage the regularization effect to provide tighter robustness and lower bounds that hold high probability. \cite{b15} challenged these attack-dependent and time costly adversarial training techniques (\cite{b1} and \cite{b92}) by introducing an algorithm called MACER, which learns robust models without using adversarial training. It outperforms \cite{b37} by a significant margin in terms of Certified top-1 accuracy on CIFAR-10 at various $l_2$ radii. On Imagenet, however, the latter gives better accuracy. 

Thus, the challenge of training a base classifier that is accurate and robust stands resolved (\cite{b37}) along with some noteworthy contributions (\cite{b15}) that prove that adversarial training is not necessary for robust training and larger average certified radii can be achieved in less training time than state-of-the-art adversarial training algorithms. We can link these results to scalability and conclude that these methods make RS somewhat scalable in terms of time.

\subsection{Improving Performance of Pre-Trained Classifiers}

Improving the performance of pre-trained classifiers by turning them from non-robust into provably robust ones is desired to maintain their confidentiality, integrity, and availability. For example, it might be beneficial to public vision API providers and users.
In their work towards achieving this, \cite{b20} apply randomized smoothing to pre-trained classifiers by attaching a denoiser before the pre-trained model. Their methodology increases the classification accuracy and can be extended for other $l_p$ cases by simply changing the noise used during training to a noise sampled from the distribution corresponding to the specific $l_p$ case. Denoised smoothing can be applied to both white and black box settings, making it the state-of-the-art approach to defending pre-trained classifiers. \cite{b57} challenge the low natural accuracy of this state-of-the-art approach and contribute further to the methodology by \cite{b20} by augmenting the joint system with a “rejector” and exploiting adaptive sample rejection, i.e., intentionally abstain from providing a prediction. Their approach has substantially improved the natural accuracy and the certification radius for computationally inexpensive denoisers. They achieve considerably better accuracy with their rejection methodology on CIFAR10 with both DnCNN-based denoising and MemNet-based denoising (\cite{b114} and \cite{b115}). \cite{b49} also improved on previous works (\cite{b20} and \cite{b57}) by introducing an improved score-based architecture for denoisers instead of classification loss, however, faced challenges concerning the novelty of their work and the possibility that the proposed method might exacerbate the weakness of randomized smoothing (i.e., slow prediction), especially in high-dimensions. Hence, while they produced higher certified accuracy of ResNet-110 on CIFAR-10 at various $l_2$ radii compared to previous works (\cite{b1} and \cite{b20}), they failed to achieve similar improvements in certified accuracy of ResNet-50 on ImageNet, indicating that their method seems to be effective for low-resolution images only. While the work so far in this domain seems somewhat promising, there remains scope for future research to further improve the certified accuracy of high-dimensional datasets with increasing $\sigma$ ($\sigma$ as defined in section I, subsection C).

\subsection{Leveraging Local Information}
The original randomized smoothing technique makes use of Monte-Carlo sampling to derive certified radii. This information is fully summarised for an input $x$ by $p_A(x)$, the probability of the top class, and does not require any other information beyond the standard deviation $\sigma$ of the Gaussian. \cite{b16} propose the use of additional gradient information, viz. $p_A'(x)$, in a technique known as second-order smoothing (SoS) to provide more robust bounds. Their approach is based on the Lipschitz property of randomized smoothing classifiers (\cite{b116}), which globally bounds the Lipschitz constant of the gradient of a smoothed network, showing that all randomized smoothing-based classifiers suffer from this curvature constraint, restricting their application and highlighting scalability issues. It also introduces Gaussian dipole smoothing as an optimisation based on SoS and Gaussian randomized smoothing to provide similar radius bounds, addressing the high inference cost issue.

\cite{b26} extend the usage of gradient information to provide a general framework for evaluating more optimal certified radii using higher-order derivatives around the input. They introduce a theoretical result for Gaussian-smoothed classifiers, which states that it is possible to achieve arbitrarily tight bounds using sufficient local information, thereby addressing the curse of dimensionality, which is a major challenge in RS.

\cite{b22} utilize a locally optimal robust classifier based on Weierstrass transformations using decision boundary information. Their model not only improves on local information methods mentioned here but also addresses the issue of the robustness-accuracy trade-off for smoothing classifiers as discussed previously. Their study is limited to binary classifiers but remains unexplored for other cases, and this is a potential area for future works to explore.

Thus, one can see that there is an improvement in certified bounds using local information, and there is a possibility to explore and apply this approach orthogonally to other optimisations for randomized smoothing. However, from the perspective of scalability, a drawback of this approach is that the computational overheads associated with calculating local information add heavily to the already expensive smoothing convolution despite the gains being limited and tightly bound (as shown in \cite{b16}). 

\subsection{Ensemble-based methods}
Randomized smoothing bounds can be improved further using ensemble-based methods. These methods work by aggregating the results of multiple classifiers, somewhat similar to how RS involves smoothing the base classifier with multiple noisy inputs and using those results to create the base classifier.

\cite{b60} empirically verify choosing ensembles as base models for RS and obtain state-of-the-art results in multiple settings. The insight around which their work revolves is that reduced variance of ensembles over the perturbations introduced in RS leads to significantly more consistent classifiers. Additionally, they introduce a data-dependent adaptive sampling scheme for RS that enables a significant decrease in sample complexity of RS for predetermined radii, thus reducing its computational overhead. The modeling approach is mathematically related to the work by \cite{b96}.
RS requires augmenting data with large amounts of noise, which leads to a drop in accuracy. Smooth-reduce, by \cite{b29}, a training-free, modified smoothing approach overcomes this by leveraging the patching of images and aggregation to improve classifier certificates. This approach is different from a simple ensemble technique, which is computationally expensive, and instead involves patch sampling. They provide theoretical guarantees for such certificates and empirically show significant improvements over other methods. They also extend this approach to videos.

The Smoothed WEighted ENsembling (SWEEN) scheme by \cite{b37} improves the robustness of RS classifiers using ensemble techniques that generally can help achieve optimal certified robustness. Theoretical analysis further proves that the optimal SWEEN model can be obtained from training under mild assumptions. Additionally, SWEEN does not limit how individual candidate classifiers are trained. The model adopts a data-dependent weighted average of neural networks to serve as the base model for smoothing. To find the optimal SWEEN model, one must minimize the surrogate loss of $g_{ens}$ over the training set and obtain appropriate weights. The major drawback of ensembling is the high execution cost during inference, for which the paper develops an adaptive prediction algorithm to reduce the prediction and certification cost of the models.

Thus, we see that ensemble-based methods for randomized smoothing have empirical and theoretical guarantees. 
Despite the prohibitive costs of ensemble training, adaptive algorithms have been shown to reduce the computational overhead. However, these algorithms are built for a particular problem and may or may not optimize all kinds of ensemble models. Overall, ensemble methods provide a satisfactory approach to improve robustness guarantees with reduced costs as compared to other similarly expensive approaches. As with other approaches, we encourage the orthogonal application of these approaches and also a study on how to reduce ensemble training overheads further, possibly by advantageously using the noise augmentation already employed for randomized smoothing.

\subsection{Input Specific Information}

RS uses a data-independent hyperparameter, $\sigma$, for the Gaussian noise $\epsilon \sim \mathcal{N}(0, \sigma^2I)$ to be convoluted with the base classifier. Even though this forms a basis of the robustness-accuracy tradeoff, it is also important to note that the issue of an optimal choice of $\sigma$ is important to improve certified robustness using TS. To deal with this, \cite{b17} propose the use of additional information, as is suggested in the curse of dimensionality section, which can be extracted from the data itself, in a scheme known as Data Dependent Randomized Smoothing (DDRS). DDRS aims to improve the choice of the hyperparameter $\sigma$ to achieve a larger certified radius, a parameter traditionally fixed in the original Randomized Smoothing (RS) method. Specifically, this is done by solving an optimisation problem using gradient ascent. \cite{b47} also develop a related sample-wise choice of $\sigma$ which they term as Insta-RS along with adaptive training optimisations for the same.

The concept of anisotropic smoothing aims to improve further on the results given by DDRS, which only certifies isotropic boundaries (\cite{b19} and \cite{b119}). Anisotropic smoothing aims to certify, as the name suggests, different certified radii along each axis, essentially acting as an axis-aligned hyperelliptic curve instead of a hypersphere when considering the $l_2$ norm. Anisotropic smoothing uses $\mathcal{N}\left(\mu,\Sigma\right)$ where $\Sigma$ is a diagonal matrix of variance generated using data-dependent techniques as the noise distribution. Their methodology is such that the certified bounds along each direction exceed and encapsulate the bounds given by isotropic data-dependent smoothing as in \cite{b17}.

In an unrelated but still data-dependent approach, \cite{b63} propose a smoothing algorithm that optimises the noise level choice (measured as the variance of the isotropic Gaussian noise) by dividing the input space into multiple regions and optimizing the noise level for each region, pre-training and fine-tuning the model using testing data. This approach allows one to create a sample-wise robust classifier that addresses the accuracy-robustness trade-off and improves the certified radii. Their approach maintains robustness in each region of the input space as per Cohen's results without affecting other regions.

\cite{b24} offers a concise overview of issues associated with a fixed global $\sigma$ value including (I) Using lower confidence bounds for class probability estimation, leading to smaller certified radii, (II) Balancing the trade-off between robustness and accuracy, as previously discussed and (III) The effect of Randomized Smoothing (RS) on the classifier's decision boundary, where bounded or convex regions contract with increasing $\sigma$ while unbounded and anti-convex regions expand. The paper then tackles these issues by introducing Input-Specific Randomized Smoothing (ISRS) which employs the Neyman-Pearson lemma in a general case. However, they also state the effect of the curse of dimensionality and propose a framework to evaluate the efficacy of data-dependent RS methods in high dimensionality.

Riemannian data-dependent randomized smoothing (RDDRS) and RANCER (\cite{b41} and \cite{b120}) further optimises the approach in anisotropic smoothing by allowing it to be robust to rotations. It does so using Riemannian optimisation to obtain $\mathcal{C}$, a full covariance matrix for the Gaussian noise. As can be seen, there is clearly a pattern on increasing the complexity of the covariance matrix for the Gaussian noise to derive better results. As motivated by \cite{b24}, however, we also note the weak theoretical foundations to justify and back the empirical results of approaches like DDRS, Insta-RS, ANCER, RDDRS, and RANCER.

Geometrically Informed Certified Robustness by \cite{b5} is an approach which exploits the underlying geometric properties of a network using information around a point, such as the certified radii of nearby points, and exploits the transitivity of certifications. \cite{b64} propose a localised smoothing algorithm for multi-output settings instead of using independent RS schemes. Their approach segments each input into various parts using different anisotropic noise distributions. In practice, they utilise soft locality, i.e. when the entire input influences the output but specific subsets have higher importance, to optimise the certified bounds and aggregate the result.

One of the issues mentioned in \cite{b24} is that of fairness where class boundary disparities affect classwise prediction accuracy and robustness. This also relates heavily to the discussion in robustness-accuracy tradeoff, where a paper highlighted the need to limit data geometry-dependent augmentation schemes for similar reasons \cite{b35}. Addressing this, cite{b23} use a modified version of the Neyman-Pearson lemma to compute the certified radius where the confidence is guaranteed to stay above a certain threshold, thus achieving a significantly better-certified radius. \cite{b39} discuss randomized smoothing with the consideration of generic smoothing distributions, which can be input-dependent and provide an approach to solve the problem of maximising certified radii using input-dependent distributions. However, the paper mainly focuses on the theoretical side with a simple yet unpractical experiment as a proof of concept.

Addressing the high cost challenge in RS, Quasi-Convexity-based Randomized Smoothing (QCRS) introduced by \cite{b62} presents an optimisation which exploits the quasiconcavity condition for the radius-variance curve in randomized smoothing, which is a weaker form of concavity. Quasiconcavity is shown to hold for most data points. Based on this, QCRS is presented as a computationally efficient approach with less overheads compared to other input-specific inference-based optimisation methods, one of the few input-specific methods to address complexity.

Noting the dearth of inference cost-reducing methods like \cite{b62}, particularly for input-specific techniques which add a significant overhead to RS, which itself is computationally expensive, we also encourage works to work towards reducing these costs, since they prohibit scalable applications of RS. We also note the weak theoretical justifications and encourage future work to analyse strongly the theoretical foundations to potentially justify and back the empirical results of approaches like DDRS, Insta-RS, ANCER, RDDRS, and RANCER. Further, the level of computational complexity needed for each subsequent optimisation has risen, without regard for optimising the scalability, nor addressing the curse of dimensionality, as mentioned in \cite{b24}. Further, even though many techniques use information beyond the Neyman-Pearson lemma, such as classifier-specific information \cite{b23}, data-dependent information and local geometric information \cite{b5}, there is not much research to say that this works towards addressing the curse of dimensionality in the input-specific case while also suffering from robustness-accuracy issues as with base RS and there should be more research into addressing this issue.

%%%%%%%%%%%%%%%%%%%%%%%%%%%%%%%%%%%%%%%%%%%%%%%%%%%%%%%%%%%%

\end{document}